\documentclass[
]{ceurart}

\usepackage{listings}
\begin{document}

\copyrightyear{2022}
\copyrightclause{Copyright for this paper by its authors.
  Use permitted under Creative Commons License Attribution 4.0
  International (CC BY 4.0).}

\conference{19th Italian Research Conference on Digital Libraries (IRCDL), February 23–-24, 2023, Bari, Italy}

\title{Disentangling Domain Ontologies}


\author[1,2]{Mayukh Bagchi}[%
orcid=0000-0002-2946-5018,
email=mayukh.bagchi@igdore.org,
]
\address[1]{DISI, University of Trento,
  Via Sommarive 9, 38123 Povo, Trento (TN), Italy.}
\address[2]{Institute for Globally Distributed Open Research and Education (IGDORE).}

\author[3]{Subhashis Das}[%
orcid=0000-0001-9663-9009,
email=subhashis.das@dcu.ie,
]
\address[3]{CeIC, ADAPT, School of Computing, Dublin City University
(DCU), Dublin 9, Ireland.}

\begin{abstract}
In this paper, we introduce and illustrate the novel phenomenon of \emph{Conceptual Entanglement} which emerges due to the \emph{representational manifoldness} immanent while incrementally modelling domain ontologies step-by-step across the following five levels: perception, labelling, semantic alignment, hierarchical modelling and intensional definition. In turn, we propose \emph{Conceptual Disentanglement}, a multi-level conceptual modelling strategy which enforces and explicates, via guiding principles, \emph{semantic bijections} with respect to each level of conceptual entanglement (across all the above five levels) paving the way for engineering conceptually disentangled domain ontologies. We also briefly argue why state-of-the-art ontology development methodologies and approaches are insufficient with respect to our characterization.
\end{abstract}


\maketitle

\section{Introduction}
\label{S1}
Consider the motivating example of the \emph{Burj Khalifa}\footnote{https://www.burjkhalifa.ae/en/}, the tallest building in the entire world. It is variously modelled as, for instance, a \emph{skyscraper} in a cadastral database, a \emph{hotel} in a tourist database, a \emph{corporate complex} in an event management database and a \emph{vertical obstruction} in an air traffic database. Such \emph{heterogeneity}, whether in the aforementioned example of \emph{Burj Khalifa}, or, in general, for any real-world entity, are \emph{ubiquitous} and are considered as instantiations of the phenomenon of \emph{semantic heterogeneity} \cite{1997-SH}. The central tenet behind such heterogeneity in conceptual modelling remains the fact that representations are fundamentally \emph{cognitive constructs} \cite{2007-RR,MB-ISKO} and are, \emph{non-trivially}, grounded in the very way in which (human) conceptualizations are causally generated from (human) experientiality \cite{2021-DDM,2021-CCQ,2022-ER}. Amongst the many socio-technical ramifications of such heterogeneity of semantic representations, we are particularly interested in the problem of harmonizing such diverse conceptual representations into a unified ontological schema which can be exploited later, for instance, to classify and integrate open heterogeneous data \cite{2018-PHD}, e.g., about the \emph{Burj Khalifa}.

We model the aforementioned instantiation of the heterogeneity of conceptual representations to be essentially that of the phenomenon of \emph{Conceptual Entanglement}, \emph{viz.}, an ordered manifoldness (`\emph{entanglement}') from the (perceptual) generation of concepts to their logical formalization, immanent across (every) conceptualizations. To that end, we outline five \emph{characteristically autonomous} yet \emph{functionally linked} levels into which \emph{Conceptual Entanglement} distributes. The first level of entanglement is generated due to the different ways in which different real-world entities can be \emph{perceived}. The second level of entanglement arises due to the different ways in which different concepts perceived can be \emph{named}. The third level of entanglement pertains to the different \emph{top-level ontological distinctions} \cite{DOLCE-I,2005-BFO} into which each of such named concepts can be semantically aligned and constrained to. The fourth level of entanglement, given the top-level semantic alignment, instantiates as the different ways in which different lexicalized concepts can be \emph{hierarchically modelled}. The fifth and the final level of entanglement occurs due to the different ways in which different concepts in the lexical hierarchy can be \emph{intensionally characterized}. 

Our solution approach has the fundamental assumption that there is \emph{``only one ‘real world’ but many different descriptions of this world depending on the aims, methodology and terminology of the observer"} \cite{2013-INSPIRE}. We commit to the thesis that such semantically heterogeneous conceptual representations across levels \emph{should constitute the foundation of ontology-driven conceptual models} \cite{2008-ODCM}. Based on the aforementioned premise, we propose \emph{Conceptual Disentanglement} as a \emph{multi-level conceptual modelling strategy} grounded in guiding \emph{normative principles}, following which the conceptual entanglements in each level (and subsequently, in its entirety) can be \emph{disentangled} to semantic bijections towards developing disentangled domain ontologies (such as in healthcare \cite{2016-Das,das2021contsonto} etc.).

The chief novelty of the proposed solution are \emph{two-fold}. Firstly, the \emph{explicit} characterization of the \emph{ordered layering} of conceptual entanglement which, though immanent, remains \emph{implicit} in mainstream semantic heterogeneity, e.g., \cite{1991-SH2,2001-SH} and ontology-driven conceptual modelling (ODCM) literature, e.g., \cite{2015-ODCM,2021-TCM}. Secondly, as a consequence of the streamlined accommodation of the layered heterogeneity of conceptualization in the form of conceptual disentanglement, the \emph{key} alignment amongst its mental model (which is language agnostic), domain ontological model (expressed, for instance, in an ontology language like OWL\footnote{https://www.w3.org/TR/owl2-overview/}) and its logical axiomatization (expressed in any flavour of first order logic, e.g., Description Logics \cite{2004-DL}) is achieved. 

The remainder of the paper is organized as follows: Section (2) describes in detail the notion of conceptual entanglement instantiated with respect to each of the aforementioned five levels. Section (3) elucidates the guiding norms based conceptual modelling strategy which can facilitate conceptual disentanglement across the five levels. Section (4) concludes the paper by briefly skimming through the research implications of conceptual (dis)entanglement.

\section{Conceptual Entanglement}
\label{S2}

\noindent It is unanimously agreed that a conceptual representation is \emph{``an abstract, simplified view of the world that we wish to represent for some purpose"} \cite{2012-AI} and is fundamentally \emph{mental} in nature \cite{2009-Ontology}. We adhere to an extended, \emph{spectrum-like} notion\footnote{See \cite{2013-OHCP} for the psycho-cognitive underpinnings of our characterization.} of conceptual representation which models concepts via a stratification across five levels \emph{viz.}:
\begin{enumerate}
    \item \emph{Perception}: dealing with the mental internalization of what is \emph{the case} in the target reality;
    \item \emph{Labelling}: dealing with how the perceived concepts should be \emph{named};
    \item \emph{Semantic Alignment}: dealing with how the named concepts should be aligned and constrained to top-level ontological distinctions;
    \item \emph{Hierarchical Modelling}: dealing with how the semantically aligned concepts should be modelled in a \emph{taxonomical hierarchy};
    \item \emph{Intensional Characterization}: dealing with how each concept in the hierarchy should be defined via its (object and data) \emph{properties}.
\end{enumerate}
Founded in the aforementioned stratification, we construe \emph{Conceptual Entanglement} to be the manifoldness between the source reality and the target conceptual representation that is \emph{innate} and \emph{unavoidable} with respect to the aforementioned levels (individually for each level and cumulatively across levels). We now exemplify and elucidate the (conceptual) entanglement occurrent in each of the levels.

\textbf{\emph{Perception (Entanglement):}} Concepts are \emph{``building blocks of thoughts"} \cite{sep-concepts} aggregated and abstracted via perceiving\footnote{Notice that (the nature of) perception, \emph{per se}, is out of scope with respect to our characterization.} what is the case in a target reality. Notice that the mental construct of (each of) the real-world referrent as well as (each of) its possible properties are \emph{concepts} in our view. We further state two universally agreed observations about concepts. Firstly, the fact that they are fundamentally \emph{mental representations} and are \emph{intensional} by nature, or, in other words, quoting \cite{2020-GGM}, \emph{``they refer to, or are about something"}. On top of that, secondly, the fact that the same \emph{referrent} or \emph{property} can be perceived differently by different \emph{agents} depending on their purpose (\emph{viewpoints}), leading to the premise that concepts are \emph{cognitive filters} \cite{2020-GGM}. We posit that the entanglement at the perception level occurs precisely due to the two aforementioned observations.

To take an example, the real-world geospatial entity \emph{Burj Khalifa} is perceived as a \emph{skyscraper} by a real-estate agent because his or her purpose might be to rent out an office suite in the building and consequently, he or she concentrates \emph{only} on cadastral-relevant properties. The same entity, however, is perceived, via a (partially overlapping) different set of properties, as a \emph{vertical obstruction} by an aviation administration agent given his or her purpose of facilitating a safe air route for all aviation users. Thus, the same entity \emph{Burj Khalifa} can be mentally internalized as different concepts based on the purpose\footnote{The state-of-the-art ontology-driven conceptual modelling approaches employed to codify purpose are not theoretically well-founded \cite{2019-CQP} and are \emph{at best approximate} in codifying different perceptions.} at hand. Similarly, the same concept of a \emph{skyscraper} or a \emph{vertical obstruction} can be equally internalized, for instance, with respect to the \emph{Shanghai Tower}, the second tallest building in the world. 

\textbf{\emph{Labelling (Entanglement):}} Given the internalization of concepts, the second level concentrates on naming or \emph{labelling} them for human and machine interaction purposes. The very act of labelling, contrary to mainstream interpretation, is \emph{non-trivial} due to the intimate way in which thought and language interoperate. We mention two highlights. The first highlight is the fact that languages are \emph{``itemized inventories"} of the target reality \cite{1954-SLC}. Secondly, the fact that each of such inventories generate a \emph{similar but not the same} lexicalization of the (same) perceived concept given their different cultural grounding \cite{2011-LT}. 

We exemplify how the aforementioned interoperation between thought and language generates the many-to-many entanglement. Firstly, we show the case within the same language, for instance, for English. The same concept of \emph{Burj Khalifa} as a \emph{hotel} can be referred to as a \emph{hotel} or an \emph{auberge} or a \emph{hospice}, each of which can further be an equally valid name for the \emph{Shanghai Tower}. Secondly, we present the case of multiple languages wherein the entanglement is two-fold. Firstly, the same concept of \emph{Burj Khalifa} can be referred to as a \emph{hotel} in English or as a \emph{Fremdenzimmer} in German, each of which can be equally valid for the \emph{Shanghai Tower}. Further, even for describing the same hotel \emph{Burj Khalifa}, English and German hoteliers might use a partially non-overlapping set of labels, each of which might describe the \emph{Shanghai Tower} equally well.

\textbf{\emph{Semantic Alignment (Entanglement):}} Given the naming of concepts, the third level concentrates on semantically aligning the lexicalized concepts to top-level (\emph{aka} foundational) ontological distinctions (themselves being grounded in distinct philosophical theories of existence). For instance, whether a specific instance of modelling a concept is semantically constrained to be an \emph{endurant} or a \emph{perdurant} \cite{DOLCE-III}, or, an \emph{independent} or a \emph{dependent} conceptual entity \cite{DOLCE-III}, or, for instance, a \emph{mental object}, a \emph{process} or an \emph{event}. This is crucial given the well-established fact that the same (perceived and subsequently named) concept can be modelled in terms of different foundational distinctions, and equally, a specific foundational distinction might semantically instantiate into different named concepts in a particular domain. 

We now briefly exemplify how the aforementioned semantic alignment raises the possibility of many-to-many entanglements at this level. For example, let us consider the case of modelling an independent versus a dependent conceptual entity. Depending on the purpose, the building \emph{Burj Khalifa} can be conceptually modelled as a \emph{building} which is an \emph{independent} conceptual entity, when the purpose for modelling is, for instance, to model a generic classification ontology of a city. The same building, however, can equally be modelled as a \emph{dependent} conceptual entity in the form of a \emph{corporate complex} when it necessarily participates in (the event of) hosting a corporate management conference or a \emph{hotel} when it participates in the (the event of) sojourn of the conference's guests. Further, the same characterization of an \emph{independent} or a \emph{dependent} conceptual entity can be semantically attributed to other buildings such as the \emph{Shanghai Tower}, thereby, generating the manifoldness.

\textbf{\emph{Hierarchical Modelling (Entanglement):}} Given the conformance of concepts to top-level semantic distinctions, the fourth level concentrates on organizing the named concepts in a \emph{taxonomical hierarchy}. It usually involves four broad phases \cite{2021-DDM,2022-ER}. Firstly, the fact that with respect to a specific depth in the taxonomic tree, there are always many different aspects which can be employed to taxonomically classify concepts of that depth into (many) subordinate concepts. Secondly, the successive application of such aspects across the entire taxonomy (with the possibility of many classificatory aspects at each depth) leads to potentially infinite classifications. Thirdly and fourthly, the many ways in which concepts can be organized horizontally across a specific taxonomic depth and vertically across a specific taxonomic path, respectively. We argue that the ordered application of the four aforementioned phases results in \emph{infinitely many} hierarchies, thereby, generating an entanglement.

For example, let us consider the case of a \emph{building}. The aspect of classifying a building can well be its \emph{purpose} (generating children like \emph{hotel, theatre} etc.) or can equally be its \emph{color} (generating children like \emph{red-colored building, blue-colored building} etc.). Further, each of these aspects can equally be applied to classify entities other than a building. Given the category \emph{hotel}, the second classification aspect, amongst infinitely many possible options, can be, for instance, \emph{number of stars}. Now, the \emph{combinatorial possibilities} of the succession of aspects also definitively determine the many modelling possibilities of sibling concepts at a single level in the hierarchy (for instance, \emph{library, secretariat, lab, sports arena} as siblings or \emph{academic buildings, non-academic buildings} as siblings). Finally, for the very same reason of the the \emph{combinatorial possibilities} of the succession of aspects, there can be \emph{infinitely many} vertical paths which can be modelled in a taxonomical hierarchy.

\textbf{\emph{Intensional Definition (Entanglement):}} Given the hierarchical modelling of concepts, the fifth and the final level concentrates on modelling the concepts at an intensional level, or, in other words, defining each individual concept in the taxonomic hierarchy via their appropriate \emph{relations} (object properties) and \emph{attributes} (data properties), thereby rendering the hierarchical model as a formal ontological schema. The manifoldness at the intensional level is generated when, each concept, in the different ontological hierarchies modelled out of the same target reality, can be differently defined via a distinct set of relations and attributes.

For example, the notion of \emph{Burj Khalifa} as a \emph{hotel} can be characterized differently via the following two sets of attributes: \emph{\{number of rooms, number of VIP suites\}} or \emph{\{year of establishment, latitude, longitude\}}. Further, each of the above attribute set can also be an equally well-defined intensional characterization for, say, the \emph{Shanghai Tower} as a \emph{hotel}. Considering also, for instance, the modelling of the relation between two concepts such as a \emph{hotel} and a \emph{conference}, there can be potentially a many-to-many entanglement, e.g.,
a \emph{hotel} can be the \emph{main venue} of a \emph{conference} and/or it can be the \emph{official entertainment junction} for a \emph{conference}. Each of such relations can further hold between several pairs of concepts  within the same domain of discourse.

There are three important observations concerning the aforementioned stratification of conceptual entanglement. Firstly, from an ablationary perspective, a specific (domain) ontology development project team, depending on their purpose, might choose to factor in all or \emph{only some} of the aforementioned levels which contribute to the conceptual entanglement. Secondly, the fact that the entanglement rooted in each of the aforementioned layers of (human) conceptualization is ultimately instantiated as \emph{semantic heterogeneity} in the (different) database representations of the same target reality. This leads to a fractured crosswalk and ultimately the lack of \emph{semantic interoperability} \cite{2019-IOP} amongst them resulting in the lack of a unified classification, integration and analysis of the target data which is heterogeneous in nature. Last but not the least, \emph{the lack of methodological support} to tackle the entanglement across the different layers of conceptualization also results in an \emph{incorrespondence} amongst its mental model (which is language-agnostic), domain ontological model (which, almost always, is expressed in a formal ontology language) and their underlying logical axiomatization (expressed in a flavour of first order logic such as \emph{description logics} \cite{2012-DL}).

\section{Conceptual Disentanglement}
\label{S3}

\noindent We propose \emph{Conceptual Disentanglement} as a multi-level conceptual modelling strategy to tackle the five-fold manifoldness of conceptual entanglement that instantiates in modelling ontologies for knowledge-based information systems (such as healthcare systems \cite{2021-NKG}) due to the nature of (human) conceptual representations as discussed in the previous sections. Conceptual disentanglement essentially refers to a set of guiding \emph{normative principles} which, if considered as best practice for each of the five levels, can enforce \emph{semantic bijections} disentangling the conceptual entanglements while at the same time, accommodating the (extent of the) heterogeneity of target reality that requires to be modelled. In other words, conceptual disentanglement can provide the conceptual modelling foundations based on which a methodology for harmonizing diverse conceptual representations into a \emph{dynamic} single ontological schema can be developed. We now elucidate and exemplify conceptual disentanglement strategy specific to each level.

\textbf{\emph{Perception (Disentanglement):}} Let us first concentrate on the norms which tackle the manifoldness instantiated due to the very nature of the \emph{Perception} level. We recommend the sequential fixation of the following:
\begin{itemize}
    \item Firstly, the \emph{target reality} should be precisely delineated with respect to their \emph{spatio-temporal} extent. In the case of a distribution of several smaller component realities, the target reality should be modelled as a disjoint union of the component realities (i.e., of component spatio-temporal extents). Notice that such a spatio-temporal delineation can be as \emph{general or specific} as possible depending upon the ontology development requirements elucidated, for instance, via Competency Questions (CQs) \cite{CQ}.
    \item Given the delineation of the target reality, the second activity should determine the concepts which requires to be modelled within the chosen target reality and the \emph{viewpoints} to be considered while modelling them.
\end{itemize}

\noindent The two aforementioned norms facilitate selection of the \emph{intended ontological commitment} of the target reality very precisely and thus avoids instances of overcommitment and undercommitment which frequently arise in every domain (see \cite{2004-OC} for related exemplifications). This, in effect, disentangles the conceptual entanglement at the perception layer to a semantic bijection. 

For example, let us consider the purpose of modelling an ontology targeted at integration of cadastral data of \emph{Burj Khalifa} and \emph{Shanghai Tower} which are the only components of our target reality. We fix the spatial extent of our target area as the disjoint union of the latitude and longitude of \emph{Burj Khalifa} (\texttt{25.1972° N, 55.2744° E}) and \emph{Shanghai Tower} (\texttt{31.2335° N, 121.5056° E}) respectively. Further, for instance, we fix the temporal extent from \texttt{2022-01-01 00:00:00} to \texttt{2022-02-01 00:00:00}. Given the spatio-temporal delineation of the target reality, we fix the viewpoint to be that agreed to by real estate agents. This entails perceiving both the \emph{Burj Khalifa} and \emph{Shanghai Tower} as a \emph{skyscraper} filtered out via relevant properties.

\textbf{\emph{Labelling (Disentanglement):}} Given the enforcement of the semantic bijection at the perception level, we lay down, as follows, the guiding norms for disentangling the conceptual entanglement occurring in the \emph{Labelling} level due to language related issues:
\begin{itemize}
    \item Fixation of the underlying natural languages(s) and the \emph{controlled vocabulary} \cite{CV}, the terminology of which has wide \emph{inter-labeller agreement} \cite{ILA} and can be used to uniquely name the concepts. International terminological standards and conventions for various domains (e.g., RESO Data Dictionary\footnote{See: https://www.reso.org/data-dictionary/} for real-estate domain) can be exploited for this purpose. Such a choice forces a semantic bijection out of the multiplicity of possible labellings in the selected language(s) on one hand, and absolves the effect of linguistically-grounded labelling conflicts such as endonym and exonym \cite{EE} on the other hand.
    \item Optionally, given the fixation of the terminology, the next step, especially key in scenarios like multilingual and cross-border data integration is to further \emph{disambiguate} the uniquely named via associating to each of such concepts a unique alphanumeric identifer such as the ones provided by general purpose knowledge graphs like Wikidata \cite{2014-WikiData}. Notice that Wikidata also provide the service of adding a new concept (with, consequently, a new alphanumeric identifier) if a certain concept is absent in it\footnote{See: https://www.wikidata.org/wiki/Help:Items}.
\end{itemize}

Let us exemplify the above norms. Consider, for the sake of simplicity of exemplification, the underlying natural language is English in case of modelling cadastral data from both the \emph{Burj Khalifa} and the \emph{Shanghai Tower}. Given the fixation of English as well as the earlier fixation of the viewpoint as a \emph{Skyscraper}, the next key activity is to uniquely label them (i.e. more non-generally than that of a `skyscraper'). If we commit to RESO as the relevant terminological (data) standard, there can, in turn, be at least three non-synonymous labels such as \emph{Residential Income}, \emph{Commercial Lease} or \emph{Commercial Sale} which can be employed to label such a concept. The key is to select the one which encodes the concept most uniquely in the face of challenging labelling conflicts such as the phenomenon of \emph{endonym-exonym}  which are key in geospatial labelling decisions. Additionally, such a label can be alphanumerically disambiguated by associating to it a unique identifier (such as from Wikidata) which will render it adept for a potential multilingual label-linkage later on.

\textbf{\emph{Semantic Alignment (Disentanglement):}} Once the concepts are encoded via a label, the next key step is to perform an \emph{ontological analysis} with respect to each of the labelled concepts (including both referrents and their relevant attributes) from the previous level, this being the guiding norm to disobfuscate the manifoldness with respect to top-level semantic alignment. Ontological analysis employs a set of metaproperties to \emph{``characterize relevant aspects of the intended meaning of the properties, classes, and relations that make up an ontology"}\cite{2002-OC}. To take an example, in the widely used OntoClean framework \cite{2002-OC,2005-OC2}, the notion of \emph{essence} and its special case of \emph{rigidity} is crucial in ascertaining the \emph{ontological stance} of geospatial entities. Similarly, analysis of concepts from the perspective of ontological \emph{identity}, \emph{unity}, \emph{endurance, perdurance} etc. facilitates, at a later stage, development of ontologically well-founded conceptual models. The cumulative target of the ontological analysis should be to determine the exact ontological nature of each labelled concept by \emph{semantically constraining them to a specific top-level ontological distinction grounded in the precise perceptual viewpoint}.

For example, the geospatial entity \emph{Burj Khalifa} should be constrained to be an \emph{independent} conceptual entity, when modelled as a \emph{building} and when the greater purpose for modelling is to model a classification ontology \cite{2018-PHD} of a city. This is precisely because a building is ontologically \emph{rigid} in the aforementioned sense (i.e., \emph{a building is always a building}). The same entity, however, should be strictly modelled as a \emph{dependent} conceptual entity in the form of a \emph{corporate complex} when it necessarily participates in (the event of) hosting a corporate management conference because the concept itself is ontologically \emph{non-rigid} (i.e., \emph{a corporate complex is only contingently so}).

\textbf{\emph{Hierarchical Modelling (Disentanglement):}}
In the context of building a disentangled taxonomical hierarchy out of ontologically analysed concepts, we ground our hierarchy modelling design in the \emph{four-step} epistemologically well-founded classification theory by Ranganathan \cite{srr67,srr89}. The first step concerns the selection of a differentiating \emph{characteristic} for classification at a single level in the hierarchy. The second step involves the \emph{succession of characteristics}, or, in other words, the selection of characteristics for classifying at each successive level in the hierarchy. The third step involves how \emph{sibling concepts} should be modelled within a single level in the hierarchy (termed \emph{array} in \cite{srr67}). The fourth and the final step concentrates on the ontological consistency of (each of) the \emph{single path} in the ontological hierarchy (termed \emph{chain} in \cite{srr67}). We now elucidate the guiding norms for each step:

\begin{itemize}
\item In the first step, we eliminate the manifoldness in the selection of the differentiating characteristic by exploiting the canons of \emph{relevance} (stating that such a characteristic should be relevant to the purpose at hand) and \emph{ascertainability} (stating that such a characteristic should be perceptually ascertainable). For example, in the case of building a classificatory ontology for buildings for government surveys, we fix \emph{legal nature of building} as the first classification characteristic.
\item The entanglement in the second step of choosing the succession of characteristics is disentangled by employing the canon of \emph{relevant succession} which enforces that the selection of successive differentiating characteristics across the depths of a taxonomy should be founded solely on purpose. For instance, the second characteristic for the classificatory ontology on buildings could be \emph{year of establishment} given the purpose is to aggregate timeseries data on real estate by a local government body.
\item The canon of \emph{exhaustiveness} is employed to eliminate the entanglement in arrays by ensuring that all the concepts at a specific depth in the taxonomic hierarchy are exhaustively classified at the next depth and thereby, additionally ensuring the exclusivity of chosen purpose-driven differentiating characteristic(s).
\item Finally, the entanglement for modelling a chain is eliminated via the canon of \emph{modulation} which ensures that there are no missing \emph{conceptual} links in any possible path of a taxonomy. For example, this canon ensures that all the paths in the classification of a domain of buildings are populated by concepts at all depths and rules out missing links (which otherwise hints at a many-to-many crossover in the succession of characteristics).
\end{itemize}

\textbf{\emph{Intensional Definition (Disentanglement):}} Given the norm-based disentanglement of the hierarchical taxonomic model, the final entanglement at the intensional definition level \cite{ISO704} is fixated by precisely determining the relations (object properties) and attributes (data properties) out of the manifoldness that ought to be encoded by the (developed) ontology for each concept in its hierarchy. 

For example, we conceptualize \emph{Burj Khalifa} as a \emph{hotel} characterized via the following set of attributes: \emph{\{number of rooms, number of VIP suites\}} given the modelling purpose of classifying and capturing the internal infrastructural details of entities in the hospitality sector. Further, for instance, we model the relation between the concepts of a \emph{hotel} and a \emph{conference} in terms of the \emph{hotel} being the \emph{main venue} of the \emph{conference} thus again ruling out many-to-many possibilities with respect to relations.

As a summative observation, notice that the conceptual disentanglement across the different levels of conceptual representation enforces a novel \emph{correspondence} amongst its mental model (which is language-agnostic), ontology (expressed in a formal ontology language) and its underlying logical axiomatization (expressed as a decidable fragment of first order logic). 

\section{Conclusive Research Implications}
\label{S4}
Given the explication of the phenomenon of \emph{Conceptual Entanglement} in domain ontologies and a multi-level conceptual modelling foundation in the form of \emph{Conceptual Disentanglement}, the next key question becomes the development of a dedicated methodology for designing as well re-engineering conceptually disentangled domain ontologies. The state-of-the-art ontology development methodologies (see \cite{METHONTOLOGY,NeOn,2021-DDM}) are not founded in (dis)entanglement and thus are not suitable to be exploited in our case. Firstly, none of the above methodologies \emph{recognize}, in an entirety, the five-layered ordered phenomenon of conceptual entanglement (perhaps, due to their difference in methodological focus). Secondly, none of the above methodologies are tailor-made for engineering conceptually disentanged classificatory ontologies \cite{2022-ONTOBRAS}, a special form of domain ontologies which are \emph{pivotal for knowledge organization and representation in digital libraries and repositories}. Moreover, the same reasons also hold for ontology development methodologies developed in the context of engineering ontologies in different domains, e.g., see \cite{2019-MS,IIM-ISI} for knowledge-graph based ontology engineering, \cite{2010-LifeScience} for life sciences, \cite{DJLIT} for chatbots, \cite{2022-LOT} for industries, \cite{2006-GOM2} for geospatial domain etc.

In conclusion, the paper introduced the novel (general) phenomenon of \emph{Conceptual Entanglement} which is \emph{unavoidable} while developing ontologically grounded conceptual models. It also proposed and exemplified the multi-level conceptual modelling strategy of \emph{Conceptual Disentanglement} as a solution to conceptual entanglement towards developing conceptually disentangled domain ontologies.

 Our ongoing work, thus, involve developing a full fledged methodology with concrete conceptual and engineering tools for developing conceptually disentangled domain ontologies.

\section*{Acknowledgement}
Supported by MF No: 222879, the EU H2020 ELITE-S MSC Grant Agreement No. 801522, SFI and the ERDF through the ADAPT CDCT Grant Number 13/RC/2106\_P2 and DAVRA Networks.

\bibliography{IRCDL}
\end{document}